\def\year{2019}\relax
\documentclass[letterpaper]{article} 
\usepackage{aaai19}  
\usepackage{times}  
\usepackage{helvet}  
\usepackage{courier}  
\usepackage{url}  
\usepackage{graphicx}  
\usepackage{float}
\usepackage{amsmath}
\usepackage{amssymb}
\usepackage{multirow}
\usepackage{booktabs}
\usepackage{array}
\usepackage{algorithm}
\usepackage{algorithmic}
\usepackage{longtable}
\usepackage{enumitem}
\usepackage{bm}
\usepackage[normalem]{ulem}
\usepackage{xcolor}

\begin{document}
%

\title{Convolutional Spatial Attention Model for Reading Comprehension with Multiple-Choice Questions}
\author{Zhipeng Chen$^\dag$, Yiming Cui$^{\dag\ddag}$\thanks{Corresponding author}, Wentao Ma$^\dag$, Shijin Wang$^\dag$, Guoping Hu$^\dag$ \\
	{$^\dag$Joint Laboratory of HIT and iFLYTEK (HFL), iFLYTEK Research, Beijing, China}\\
	{$^\ddag$Research Center for Social Computing and Information Retrieval (SCIR),}\\
	{Harbin Institute of Technology, Harbin, China}\\
	{\tt\{zpchen,ymcui,wtma,sjwang3,gphu\}@iflytek.com}\\ 
}

\maketitle
\begin{abstract}
Machine Reading Comprehension (MRC) with multiple- choice questions requires the machine to read given passage and select the correct answer among several candidates. In this paper, we propose a novel approach called Convolutional Spatial Attention (CSA) model which can better handle the MRC with multiple-choice questions. The proposed model could fully extract the mutual information among the passage, question, and the candidates, to form the enriched representations. Furthermore, to merge various attention results, we propose to use convolutional operation to dynamically summarize the attention values within the different size of regions. Experimental results show that the proposed model could give substantial improvements over various state-of- the-art systems on both RACE and SemEval-2018 Task11 datasets.
\end{abstract}

\section{Introduction}\label{introduction}
Owing to the rapid release of various large-scale datasets, Machine Reading Comprehension (MRC) has become enormously popular in Natural Language Processing.
For example, cloze-style MRC (such as CNN/DailyMail \cite{Hermann2015Teaching}, Children's Book Test (CBT) \cite{hill2015goldilocks}), span-extraction MRC (such as SQuAD \cite{Rajpurkar2016SQuAD}), and multiple-choice MRC (such as MCTest \cite{Richardson2013MCTest},  RACE \cite{lai2017race}, SemEval-2018 Task11 \cite{Ostermann2018SemEval}).

In this paper, we mainly focus on solving the reading comprehension with multiple-choice questions. At the beginning of the reading comprehension study, this type of reading comprehension task was not that popular because there is no large-scale dataset available and thus we cannot apply neural network approaches to solve them.
To bring more challenges to reading comprehension task and mitigate the absence of large-scale multi-choice reading comprehension dataset, \citeauthor{lai2017race}\shortcite{lai2017race} propose a new dataset called RACE. 
Compared to the earlier MCTest \cite{Richardson2013MCTest}, the RACE dataset is made from the English examinations for Chinese middle and high school students, consisting near 100,000 questions generated by human experts, and is far more challenging than the MCTest.

In this paper, we propose a novel model called Convolutional Spatial Attention (CSA) to fully utilize the hierarchical attention information for reading comprehension with multiple-choice questions. 
The proposed model first encode the passage, question, and candidates into word representations which are enhanced by the additional POS-tag and matching features. 
Then we concentrate on enriching the representation of the candidates by incorporating the passage, question information, and further calculate the attentions between the passage, question, and candidates, forming the spatial attentions.
To further extract the representative features in the spatial attentions, we propose to use convolutional neural network to dynamically conclude adjacent regions with different window size.
We mainly test our CSA model on two multiple-choice reading comprehension datasets: RACE and SemEval-2018 Task11, and our model achieves state-of-the-art performances on both of them. The examples of each dataset are given in Figure \ref{fig:1}.
The main contributions of our paper can be summarized as follows.
\begin{itemize}
	\item We focus on modeling different semantic aspects of the candidates, by integrating the passage and question information, forming the 3D spatial attention among the passage, question, and candidates.
	\item We propose a Convolutional Spatial Attention (CSA) mechanism to dynamically extract representative features from the spatial attentions.
	\item The proposed model gives substantial improvements over various state-of-the-art systems on both RACE and SemEval-2018 Task11 datasets, showing its generalization and extensibility to other NLP tasks.
\end{itemize}
\begin{figure*}[htbp]
	\centering 
	\small
	\begin{tabular}{p{1\columnwidth}p{1\columnwidth}}
		\toprule
		\bf RACE & \bf SemEval-2018 Task11 \\ 
		\midrule
		\bf Passage & \bf Passage\\ 
		Is it important to have breakfast every day? A short time ago, a test was given in the United States. People of different ages, from 12 to 83, were asked to have a test. During the test, these people were given all kinds of breakfast, and sometimes they got no breakfast at all.
		...
		&
		I was thirsty so I decided to make a cup of tea. I looked through my box of teas and rifled through the assorted flavors. I settled on Earl Gray, which is a black tea flavored with bergamot orange. I filled the kettle with water and placed it on the stove, turning on the burner so that it would heat up and begin boiling.
		...
		\\
		\midrule
		\bf Question &\bf Question\\
		What do the results show?&Why did they use a kettle?\\
		\midrule
		\bf Candidates  &\bf Candidates \\
		{\bf A \em They show that breakfast has affected on work and study.}&{\bf A}    to drink from \\
		{\bf B}   Breakfast has little to do with a person's work. &{\bf B   \em to boil water} \\ 
		{\bf C}   A person will work better if he only has fruit and milk.\\
		{\bf D}   They show that girl students should have less for breakfast.\\
		\bottomrule
	\end{tabular}
	\caption{Example of RACE and SemEval-2018 Task11 dataset. The correct answer is depicted in bold face.}\label{fig:1} 
\end{figure*}

\section{Related Works}\label{related-works}
Massive progress has been made on machine reading comprehension field in recent years.
The booming of the MRC can trace back to the release of the large-scale datasets, such as CNN/DailyMail \cite{Hermann2015Teaching} and CBT  \cite{hill2015goldilocks}).
After the release of these datasets, various neural network approaches \cite{Chen2016A,Kadlec2016Text,cui-etal-2017,Dhingra2017Gated} have been proposed and become fundamental components in the future studies. 
Another representative dataset is SQuAD \cite{Rajpurkar2016SQuAD}, which was difficult than the cloze-style reading comprehension and requires the machine to generate a span in the passage to answer the questions.
With rapid progress on designing effective neural network models \cite{Xiong2016Dynamic,Seo2016Bidirectional,Wang2017Gated,Hu2017Reinforced}, recent works on this datasets have surpassed the average human performance, such as QANet \cite{qanet2018} etc.

However, current machine reading comprehension models are still struggling with solving the questions that need reasoning over multiple sentences or even passage. 
To solve the reading comprehension with multiple-choice questions, various approaches have been proposed, and most of them are focusing on designing effective attentions or persuing enriched representations for prediction.
When releasing the RACE dataset, \citeauthor{lai2017race}\shortcite{lai2017race} also adopted and modified two models of the cloze-style reading comprehension: Gated Attention Reader \cite{Dhingra2017Gated} and Stanford Attentive Reader \cite{Chen2016A}.
However, experimental results show that these models are not capable of this task. 
\citeauthor{parikh2018eliminet}\shortcite{parikh2018eliminet} introduced ElimiNet which use a combination of elimination and selection to get refined representation of the candidates.
\citeauthor{Xu2017Towards}\shortcite{Xu2017Towards} proposed the Dynamic Fusion Networks (DFN), which uses multi-hop reasoning mechanism for this task. 
\citeauthor{AAAI1816331}\shortcite{AAAI1816331} proposed the Hierarchical Attention Flow model, which leverage candidate options to model the interactions among passage, questions, and candidates.

Though various efforts have been made, we believe there is still a large room for designing effective neural networks for better characterizing the relations between the passage, questions, and candidates.
To this end, we summarize the main differences between our model and existing models for this task into three aspects.
First, we calculate various representations of the candidate to better characterize it for prediction.
Second, when calculating attention, we apply additional trainable weights to dynamically adjust the attention values, which is more flexible.
Third, unlike the previous works only utilize the final-level hierarchical attentions, we propose to use every attention in each hierarchy, and use convolutional neural network to capture features for predicting the answer.

\section{Convolutional Spatial Attention Model}\label{iaoa}
\subsection{Task Definition}
The RACE dataset \cite{lai2017race} for reading comprehension with multiple-choice questions was proposed, which consists of 28,000+ passages and near 100,000 questions generated by human experts for the English examinations of Chinese middle and high school students.
Different from the earlier MCTest dataset \cite{Richardson2013MCTest}, the RACE dataset is significantly larger, and thus we can apply deep learning approaches for this task. 
As all the questions and choices are generated by human experts, RACE dataset provides more comprehensive and realistic evaluation on machine reading comprehension than the other popular datasets such as CNN/DailyMail, SQuAD datasets, whose answer should appear in context. 
Also according to the analysis by \citeauthor{lai2017race}\shortcite{lai2017race}, a large portion of the questions in RACE need reasoning over various clues, which makes it more challenging and suitable to evaluate the ability of the reading comprehension systems. SemEval-2018 Task11 is closely the same with the RACE dataset but with two candidates and small size. The examples of each dataset are given in Figure \ref{fig:1}.

\subsection{The Model}
\begin{figure*}
	\centering \includegraphics [width=0.98\textwidth]{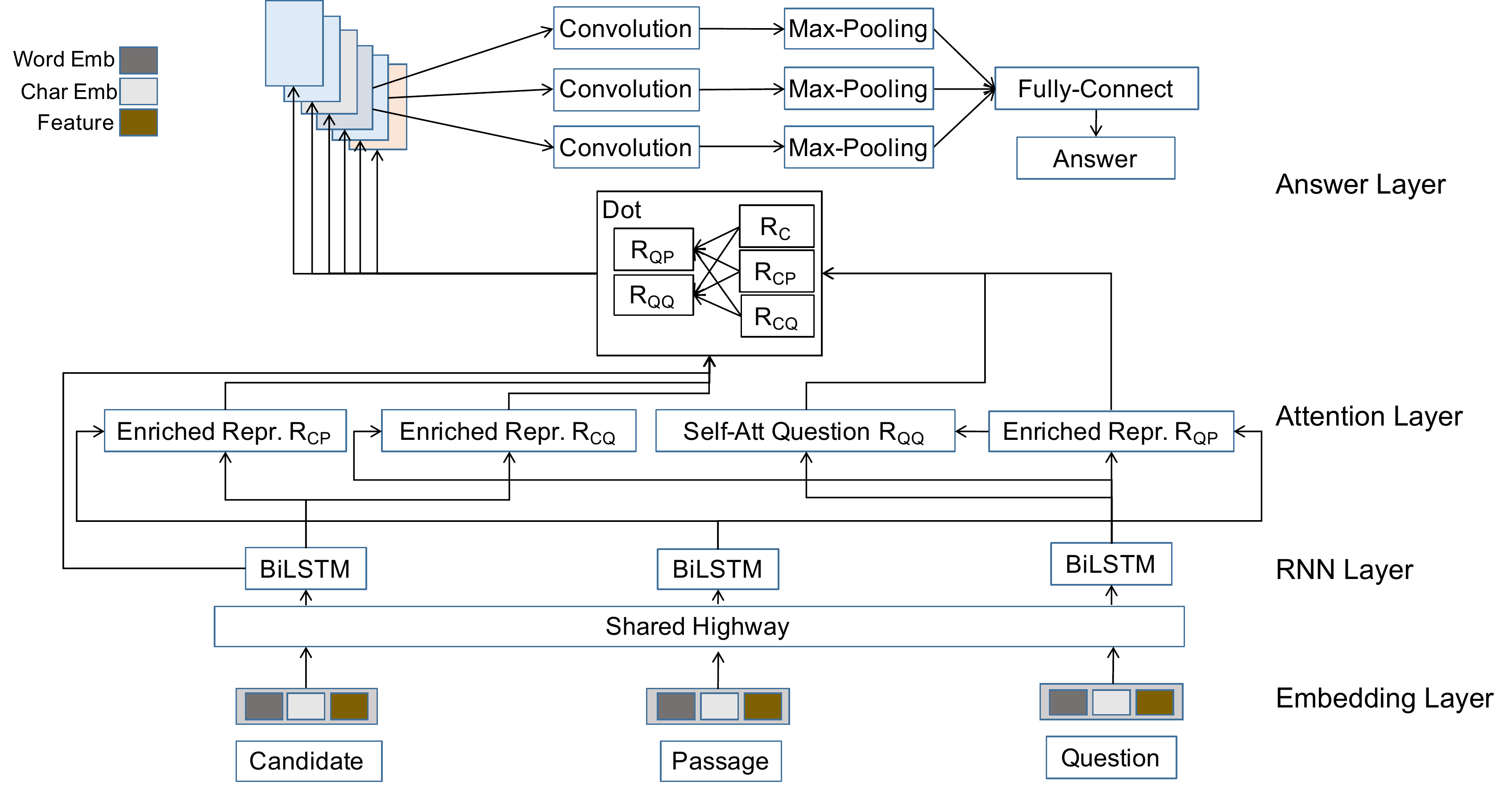}
	\caption{Main neural architecture of the Convolutional Spatial Attention (CSA) model. }\label{fig:2} 
\end{figure*}

In this section, we will give a detailed description on the proposed model. 
The main neural architecture of our model is depicted in Figure \ref{fig:2}.
Throughout this section, we will use $P$ for representing the passage, $Q$ for the question, $C$ for the candidates.
Note that, as the operations on each candidate are the same, for simplicity, we only take one of the candidates for illustration.

\subsection{Word Representation}
We transform each word in the passage, question, and candidates into continuous representations.
In this paper, there are three components in the embedding layer, which can be listed as follows.
\begin{itemize}[leftmargin=*]
	\item {\bf Word Embedding $E_{word}$}: We use traditional pre-trained GloVe embedding for initialization \cite{Pennington2014Glove} and keep fixed during the training process.
	\item {\bf ELMo Embedding $E_{elmo}$}: For this part, we use pre-trained ELMo \cite{N18-1202} embedding. 
	\item {\bf Feature Embedding $E_{feat}$}: We also utilize three additional features to enhance the word representations.
	\begin{enumerate}
		\item {\bf POS-tag Embedding $E_{pos}$}: We use NLTK \cite{Bird2004NLTK} for part-of-speech tagging for each word. Similar to traditional word embeddings, we assign different trainable vectors for each part-of-speech tag.
		\item {\bf Word Matching $F_{match}$}: Take the text as an example, if the word in text also appears in question or candidate, we set the value as one, otherwise set it as zero. In this way, we can also add this feature to the question and candidate.
		\item {\bf Fuzzy Word Matching $F_{fuzzy}$}: Similar to the word matching feature, but we loosen the matching criteria as partial matching. For example, we regard `teacher' and `teach' as fuzzy matching, because the string `teacher' is partially matched by `teach'.
	\end{enumerate}
\end{itemize}

We concatenate three embedding components to form the final word representations for the text $E_P \in {\mathbb{R}}^{|P| \times e}$, question $E_Q \in {\mathbb{R}}^{|Q| \times e}$, and candidates $E_C \in {\mathbb{R}}^{|C| \times e}$, where $|P|, |Q|, |C|$ are the length of the passage, question, and candidates, $e$ is the final embedding size (including all three components).
\begin{gather}
E = [E_{word};E_{elmo};E_{feat}]  \\
E_{feat} = [E_{pos}; F_{match}; F_{fuzzy}]
\end{gather}

After obtaining embedding representations, we further feed each word embedding into a shared highway network \cite{srivastava2015highway} with $\tanh$ output activation (denoted as $\sigma$).
In this paper, we apply two consecutive highway networks with shared weights.
Then we use Bi-Directional LSTM \cite{Graves2005Framewise} to model the contextual information, forming
$H_P \in {\mathbb{R}}^{|P| \times h}$, $H_Q \in {\mathbb{R}}^{|Q| \times h}$, and $H_C \in {\mathbb{R}}^{|C| \times h}$ ($h$ is hidden size of Bi-LSTM).
Note that, we use different Bi-LSTMs for the passage, question, and candidates.
\begin{gather}
\widetilde{H} = \sigma(\text{2-Highway}(E)) \\
H =\text{Bi-LSTM}(\widetilde{H})
\end{gather}

\subsection{Enriched Representation}
Calculating attention and generating enriched representation play very important roles in machine reading comprehension.
In our model, we will calculate various types of enriched representations for better characterizing the candidate and question, which are the essential components in this task.
The procedure for generating enriched representation is illustrated in Algorithm \ref{alg:mr}.

For example, we wish to embed the passage information into the candidate representation to better aware the relevant part in the passage and obtain the passage-aware candidate representation $R_{CP}$. 
According to Algorithm \ref{alg:mr}, $R_{CP}$ can be generated as follows.
\begin{itemize}[leftmargin=*]
	\item {\bf [Line 1]} Given the Bi-LSTM representations of passage $H_P$ and candidate $H_C$, we first calculate the attention matrix where each element indicate the matching information between them.  
	In this paper, we adopt the attention mechanism used in FusionNet \cite{huang2017fusionnet}. where two representations are transformed by individual fully-connected layer with an output activation $f$. Also, a trainable diagonal weight matrix $D$ is applied. The activation function $f$ is defined as RELU throughout this paper. 
	\item {\bf [Line 2]} Then we apply an element-wise weight matrix $W$ to the attention matrix $M'$. This is designed to let the model flexibly adjust the attention values.
	\item {\bf [Line 3]} We apply {\em softmax} to the weighted attention matrix $M$ to the last dimension of it, which calculates the passage-level attention vector w.r.t. each candidate word.
	\item {\bf [Line 4]} After obtaining the normalized attention matrix $M_{att}$, we make a dot product between the $M_
	{att}$ and the passage $H_P$ to extract candidate-related passage.
	\item {\bf [Line 5]} Finally, we concatenate candidate-related passage $Y_{CP}'$ and Bi-LSTM candidate representation $H_C$, and feed them to a Bi-LSTM to fully integrate passage information into the candidate representations.
\end{itemize}

\renewcommand{\algorithmicrequire}{\textbf{Input:}} 
\renewcommand{\algorithmicensure}{\textbf{Output:}}
\begin{algorithm}[tbp]   
	\caption{Enriched Representation.}   
	\label{alg:mr}   
	\begin{algorithmic}[1] 
		\REQUIRE ~~\\
		Time-Distributed representation $X_1$ \\
		Time-Distributed representation $X_2$ \\
		\renewcommand{\algorithmicrequire}{\textbf{Initialize:}} 
		\REQUIRE ~~\\
		Random weight matrix $W_1 \in {\mathbb{R}}^{h \times h_{att}}$ \\
		Random weight matrix $W_2 \in {\mathbb{R}}^{h \times h_{att}}$ \\
		Diagonal weight matrix $D \in {\mathbb{R}}^{h_{att} \times h_{att}}$  \\
		All-one weight matrix $W \in {\mathbb{R}}^{|X_1| \times |X_2|}$ \\
		\ENSURE $X_2$-aware $X_1$ representation $Y$
		\STATE Calculate attention matrix $M' \in \mathbb{R}^{|X_1| \times |X_2|}$: \\ $M' = f(W_1 X^1)^T \cdot D \cdot f(W_2 X^2)$
		\STATE Apply element-wise weight: $M = M' \odot W$
		\STATE Apply softmax function to the last dimension of $M$: $M_{att} = softmax(M)$
		\STATE Calculate raw representation $Y' \in {\mathbb{R}}^{|X_2| \times h}$: \\ $Y' = {M_{att}}^T \cdot X_1$
		\STATE Concatenate raw representation $Y'$ and raw input $X_1$, then apply Bi-LSTM: \\$Y = \text{Bi-LSTM}([X_1; Y'])$	
		\RETURN $Y$
	\end{algorithmic}  
\end{algorithm}

By applying the proposed algorithm $g(X_1, X_2)$, we can calculate the passage-aware and question-aware candidate representation $R_{CP}$ and $R_{CQ}$.
Besides, as the question information is also important, we also calculate passage-aware question representation $R_{QP}$ and self-attended question representation $R_{self\text{-}Q}$.
Note that, we use Bi-LSTM output $B_Q$ and passage-aware question representation $R_{QP}$ to obtain the (almost-)self-attended question representation, which combines different levels of representations.
\begin{align}
R_{CQ} &= g(B_C, B_Q) \\
R_{CP} &= g(B_C, B_P) \\
R_{QP} &= g(B_Q, B_P) \\
R_{self\text{-}Q} &= g(B_Q, R_{QP})
\end{align}

\subsection{Convolutional Spatial Attention}
With previously generated representations, we can calculate the matching matrix to measure the similarity between them.
In this paper, we adopt simple dot product to obtain the matching matrix.
As the candidate information is important to answer the question, firstly we calculate the matching matrix using various candidate representations to the self-attended question representation $R_{self\text{-}Q}$.
The motivation is to use the question information as the key to extracting candidate information in different levels.
We use question-aware candidate representation $R_{CQ}$, passage-aware candidate representation $R_{CP}$, and candidate Bi-LSTM representation $H_C$, as shown below.
\begin{align}
M_{11} &= R_{CQ} \cdot R_{self\text{-}Q} \label{eq-m11} \\
M_{12} &= R_{CP} \cdot R_{self\text{-}Q} \label{eq-m12} \\
M_{13} &= H_{C} \cdot R_{self\text{-}Q} \label{eq-m13}
\end{align}

In a similar way, we can also replace the self-attended question representation $R_{self\text{-}Q}$ with passage-aware question representation $R_{QP}$ in Equation \ref{eq-m11}, \ref{eq-m12}, \ref{eq-m13}, to obtain $M_{21}, M_{22}, M_{23}$.
Then we concatenate all matrices on the channel dimension to form a {\em Spatial Attention Cube} $M \in \mathbb{R}^{6 \times |C| \times |Q|}$, which is similar to an `image' with 6 channels.
\begin{gather}
M = [M_{11}; M_{12}; M_{13}; M_{21}; M_{22}; M_{23}]
\end{gather}

In order to extract high-level features inside the spatial attention cube, we use CNN-MaxPooling operation to dynamically conclude adjacent attention information, which is similar to the traditional operation on the image.
Formally, we first apply convolutional operation on the $M$ to summarize different length of adjacent elements
We adopt three convolutional kernels: 5, 10, and 15, along with the dimension of the question length.

After convolution operation, we could get three features maps w.r.t. different convolutional kernels.
The procedure can be illustrated as the following equations.
\begin{align}
O_1 &= \text{Max-Pooling}_{1 \times 3}\{CNN_{1 \times 5}(M)\} \\
O_2 &= \text{Max-Pooling}_{1 \times 2}\{CNN_{1 \times 10}(M)\} \\
O_3 &= \text{Max-Pooling}_{1 \times 1}\{CNN_{1 \times 15}(M)\}
\end{align}

In this way, we used CNN-MaxPooling operation to obtain three feature vectors $O_1, O_2, O_3$ by using different convolutional kernels and max-pooling intervals.

\subsection{Final Prediction}
After obtaining three feature vectors, we flatten, concatenate, and feed them into a fully-connected layer to get a scalar value denoting the possibility of being the correct answer.
Recall that, we have several candidates for a given question, so we will get $N$ candidate scores in this stage.
We apply softmax function to these scores to obtain the final probability distributions over the candidates.
\begin{gather}
s_i =  {\bf w^T} \cdot [O_1; O_2; O_3]  \\
Pr(A | P, Q, C) = softmax([s_1; ...; s_N])
\end{gather}

To train our model, we use traditional cross entropy loss to minimize the gap between the prediction and the ground truth.

\begin{table*}[ht!]
	\begin{center}
		\begin{tabular}{lccc}
			\toprule 
			\bf Model & \bf RACE-M & \bf RACE-H & \bf RACE \\
			\midrule
			Sliding Window \cite{lai2017race} & 37.3& 30.4& 32.2 \\
			Stanford AR \cite{lai2017race} & 44.2 &43.0 & 43.3 \\
			GA Reader \cite{lai2017race}  & 43.7& 44.2& 44.1 \\
			ElimiNet \cite{parikh2018eliminet}& N/A& N/A &44.5 \\
			Hierarchical Attention Flow \cite{AAAI1816331} &45.0& {\em 46.4} & 46.0\\
			Dynamic Fusion Network \cite{Xu2017Towards} & {\em 51.5} &45.7& {\em 47.4} \\
			\midrule
			CSA Model (single model) & 51.0 & 47.3 & 48.4 \\
			CSA Model + ELMo (single model) & \bf 52.2 & \bf 50.3 & \bf 50.9 \\
			\midrule
			\midrule
			GA Reader (6-ensemble) & - & - & 45.9 \\
			ElimiNet (6-ensemble) & - & - & 46.5 \\
			GA + ElimiNet (12-ensemble) & - & - & 47.2 \\
			Dynamic Fusion Network (9-ensemble) & {\em 55.6} & {\em 49.4} & {\em 51.2} \\
			\midrule		
			CSA Model (7-ensemble)  & 55.2 & 52.4 & 53.2 \\
			CSA Model + ELMo (9-ensemble)  & \bf56.8 & \bf 54.8 & \bf 55.0 \\
			\bottomrule
		\end{tabular}
	\end{center}
	\caption{\label{results-race}Experimental results on RACE. The best previous results are in italics, and overall best results are in bold face.}
\end{table*}

\section{Experiments}\label{experiments}        
\subsection{Experimental Setups}
To evaluate our system, we carried out experiments on the following two public datasets.
\begin{itemize}
	\item {\bf RACE}: English examinations for Chinese middle and high school students. The questions are generated by human experts, which has four candidates for each question. The test set consists of 4,934 instances, where RACE-M (middle school) has 1,436 instances, and RACE-H (high school) for 3,498 instances.
	\item {\bf SemEval-2018 Task11}: The dataset provided by the SemEval-2018 Task11 organizer, which mainly focus on solving commonsense reading comprehension. Each question has two candidates to choose from.
\end{itemize}

The data are tokenized and lower-cased by using Natural Language Toolkit (NLTK) \cite{Bird2004NLTK}, and all punctuations are removed.
The main hyper-parameters of our model are listed in Table \ref{tab1}.
Note that, except for the candidate numbers, all hyper-parameters are identical among two datasets.
The word embeddings are initialized by the pre-trained GloVe word vectors (Common Crawl, 6B tokens, 100-dimension) \cite{Pennington2014Glove}, and keep fixed during training. 
The words that do not appear in the pre-trained word vectors are set to the {\tt unk} token and initialized accordingly.
We use Adam \cite{DBLP:journals/corr/KingmaB14} for weight optimizations with an initial learning rate of 0.001.
In order to prevent overfitting, we apply dropout of 0.35 to all the representation layers.
The models are built on Keras platform \cite{chollet2015keras} with Tensorflow backend \cite{abadi2016tensorflow}.

\begin{table}[tbp]
	\begin{center}
		\begin{tabular}{llc}
			\toprule 
			\bf Symbol & \bf Descriptions & \bf Size \\
			\midrule
			$|P|$  & Passage max length & 300 \\  
			$|Q|$ & Question max length & 20 \\  
			$|C|$  & Candidate max length & 10 \\ 
			$e$ & Word embedding & 200 \\
			$h$ & Bi-LSTM hidden size & 250 \\
			$h_{att}$ & Attention hidden size&80 \\
			$es$ & ELMo embedding size &1024 \\
			$p$ & POS-tag embedding &16 \\
			\bottomrule
		\end{tabular}
	\end{center}
	\caption{\label{tab1}Hyper-parameter settings. }
\end{table}

\subsection{Overall Results}

{\bf RACE.}
The experimental results are shown in Table \ref{results-race}.
As we can see that, our CSA model shows significant improvements over various state-of-the-art systems by a large margin.
We also compared our model to the recent unpublished work DFN, while our model gives an absolute gain of 1.0\% in the overall test set, and further gains can be obtained by incorporating ELMo, demonstrating the effectiveness of our proposed model.
Also, when compared to the Hierarchical Attention Flow model \cite{AAAI1816331}, our CSA model show substantial improvements, indicating that utilizing the attentions in each hierarchy and using convolutional neural network to extract the most representative features from the spatial attentions are useful in this task.

When it comes to the ensemble results, though we only use 7 models in ensemble with majority voting approach, the overall results show an absolute gain of 2\% over the previous state-of-the-art result by DFN.
Also we observed that our model shows slightly worse result in RACE-M but significantly better in RACE-H regardless of single model or ensemble.
These results indicate that our model is more capable of solving difficult question (high school).
While on the contrary, it may hurt the performance on the relatively easier question (middle school).
We will give a detailed analysis for illustrating this phenomenon in the next section.

{\noindent \bf SemEval-2018 Task11.}
The distributions of question type is quite different between the RACE and SemEval-2018 Task11 datasets.
To test if our model could generalize to other reading comprehension dataset, we also carried out experiments on the very recent SemEval-2018 Task11 dataset.
The results are shown in Table \ref{results-semeval}.
As we can see that, our CSA model could give moderate improvements over the top-ranked SemEval systems in both single model and ensemble, and set up new state-of-the-art performance on this task, demonstrating the proposed model is powerful and showing its potential possibility to generalize to other NLP tasks.
Note that, we directly train the model using the hyper-parameter settings of the RACE experiments without finding other hyper-parameter combinations, indicating further improvements may be obtained through fine-tuning. 

\begin{table}[tp]
	\begin{center}
		\begin{tabular}{lcc}
			\toprule 
			\bf Model & \bf Dev & 
			\bf Test \\
			\midrule
			HMA \cite{Chen2018HFL} & \bf 84.48 &
			 80.94 \\
			TriAN \cite{DBLP:journals/corr/abs-1803-00191}& 83.84 &
			 81.94 \\
			\midrule
			CSA Model  (single model) ~~~~~ & 83.63 & 82.20 \\
			CSA Model + ELMo (single model) ~~~~~ & 83.84 & \bf 83.27 \\
			\midrule
			\midrule
			TriAN (ensemble) & 85.27 &
			83.95 \\
			HMA (ensemble) &\bf 86.46 & 
			84.13 \\
			\midrule
			CSA Model (ensemble) & 84.05 & 84.34 \\
			CSA Model + ELMo (ensemble) & 85.05 & \bf 85.23 \\
			\bottomrule
		\end{tabular}
	\end{center}
	\caption{\label{results-semeval} Experimental results on SemEval-2018 Task11. The two top-ranked systems are listed as baselines.}
\end{table}

\subsection{Ablation Study}
We also carried out model ablations to further demonstrate the effectiveness of the proposed approaches.
The results are shown in Table \ref{results-ablation}.

\begin{table}[htbp]
	\begin{center}
		\begin{tabular}{lc}
			\toprule 
			\bf Model &\bf RACE \\
			\midrule
			CSA Model & 48.52 \\
			w/o attention weight &  48.18 \\
			w/o enriched representation & 47.52 \\
			w/o convolutional spatial attention &  47.30 \\
			\midrule
			CSA Model + ELMo & 50.89 \\
			w/o attention weight &  49.49 \\
			w/o enriched representation & 49.78 \\
			w/o convolutional spatial attention &  48.47 \\
			\bottomrule
		\end{tabular}													
	\end{center}
	\caption{\label{results-ablation}Ablations on several model components. }
\end{table}

As shown in the model description, we adopt the attention mechanism used in FusionNet \cite{huang2017fusionnet}, and applied an element-wise weight to the attention matrix.
By applying additional trainable weights to the attention matrix could give moderate improvements suggesting that these weights are useful in dynamically adjusting the attention values.
When we remove all the enriched candidate representations, there is an absolute drop of 1.0\%. This suggests that it is necessary to incorporate various information (such as the passage and question) into the candidate representation.
We also removed our convolutional spatial attention mechanism in the model. 
The results show a significant drop in performance by 1.22\%, indicating that the proposed convolutional spatial attention is effective in extracting most representative values among the various attention matrices. The same results drop is shown by incorporating ELMo.
Further more, whether we use ELMo or not, our convolutional spatial attention has improved significantly. That also proved that our model can get father improvement after some big data trained toolkit like ELMo or some other transformer.

The details of the ablations are shown below.\\
\begin{itemize}
\item {\bf w/o attention weight}: We change $M=M'\odot W$( in Alg \ref{alg:mr} line 2) to $M=M'$, where $W$ is the attention weight.
\item {\bf w/o enriched representation}:  We remove interaction of candidate-to-question ($R_{CQ}$ in Figure 2 ) and candidate-to-passage ($R_{CP}$ in Figure 2) and only use candidate's LSTM output $R_C$.
\item {\bf w/o convolutional spatial attention}: We use two fully-connected layer to transform the matching matrix to a score. The first fully-connected layer squeeze the matching matrix to a vector along the question length dimension. The second fully-connected layer squeeze the vector to a scalar score. As every candidate has six matching matrices, so we can get six scores. Finally, we add the six scores as the final prediction.
\end{itemize}

\section{Analysis and Discussion}\label{analysis}
\subsection{Quantitative Analysis}
\begin{figure}[tb]
	\centering
	\includegraphics [width=0.45\textwidth]{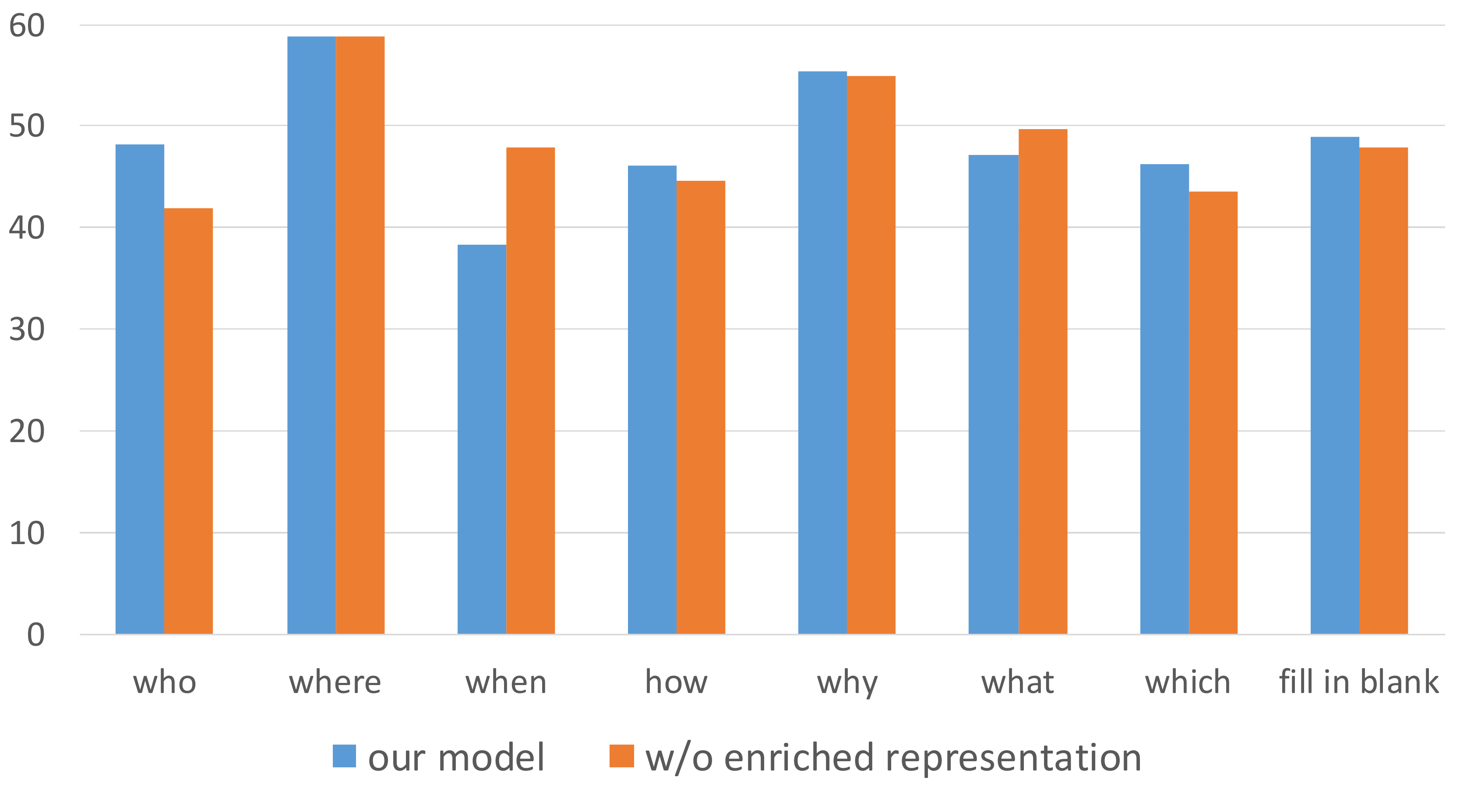} \\
	\includegraphics [width=0.45\textwidth]{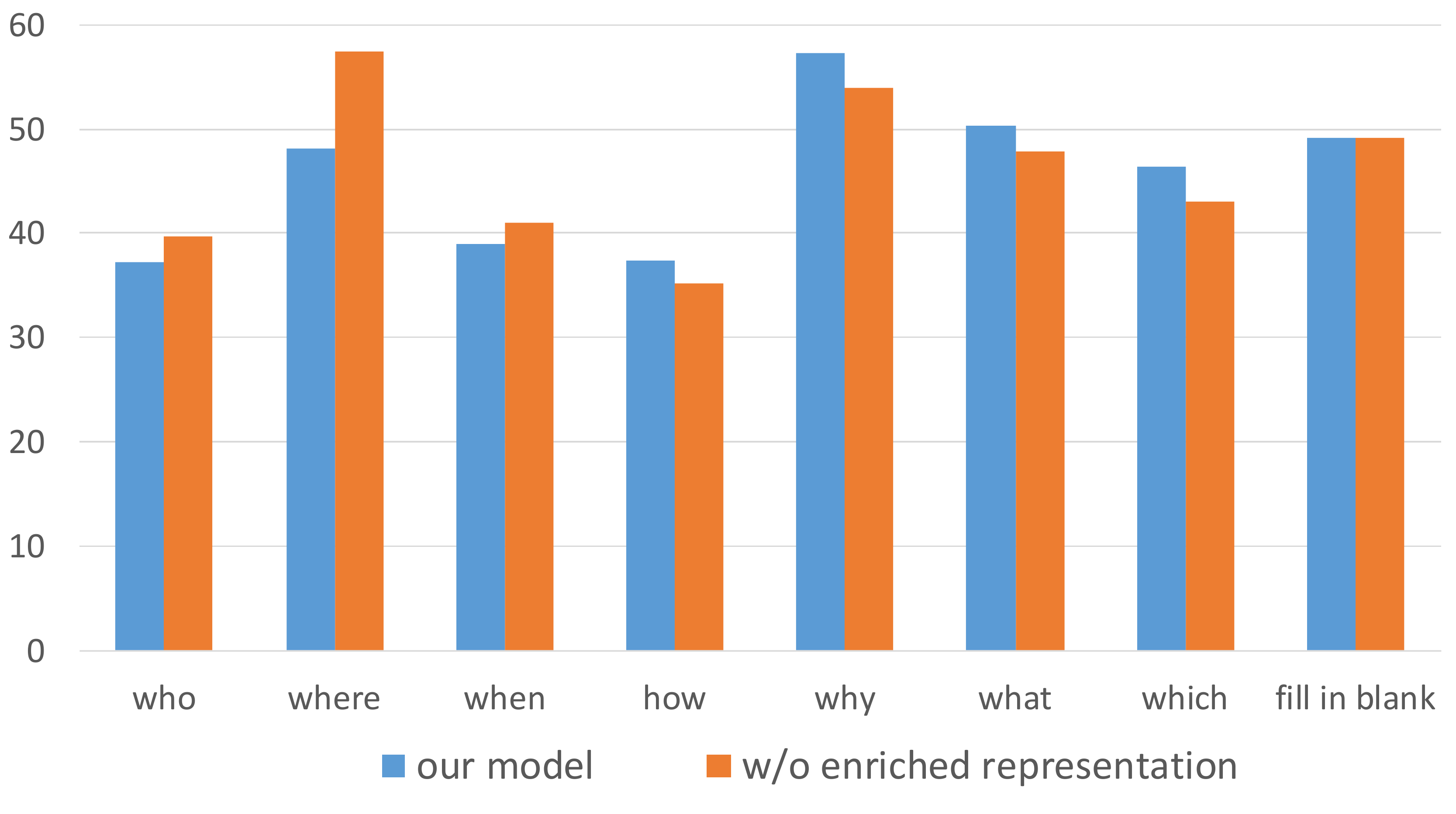}
	\caption{Quantitative analysis on different type of questions in RACE development and test set.}\label{fig-quantitative} 
\end{figure}

In CSA model, we adopt various enriched representation for better characterizing candidate information.
We compare the performance between the original CSA model and discard enriched candidate representation, as shown in Figure \ref{fig-quantitative}.

As we can see that our model yields significant improvements on the question type `how' and `why', which are relatively difficult than the other type of questions requiring high-level reasoning within the context, demonstrating that our CSA model performs better on the relatively sophisticated questions.
On the contrary, we find that our CSA model shows relatively inferior performance on the question type `who', `where', and `when', which are often answered by a single word or entity. 
This phenomenon suggests that further efforts should be made on balancing the word-level attention and highly abstracted attention.
These results also explain why our CSA model shows significant improvements on RACE-H subset while giving a relatively inferior performance on RACE-M subset.
As the sample number of RACE-H subset is three times the size of RACE-M, the overall performance of our model still shows significant improvements over various state-of-the-art systems.

\subsection{Case Study}
We also randomly sampled one hundred question from RACE test set, and classified them into three categories according to its difficulty. 
\begin{itemize}
\item The first level is the question that can be answered by matching a few words in the passage. If the model finds the right place in the text, the answer is easy to find. With the attention mechanism, the neural model is especially good at solving this kind of question.
\item The second level is the question that can be answered by using a few sentences without reasoning, which is relatively difficult than the first level.
\item The third level is the question that needed to comprehensive reasoning via multiple clues in the passage. This kind of question is more complicated. To answer this question, the intention of the questioner must be captured.
\end{itemize}
The results are shown in Table \ref{tab5}.
As we can see that the current model shows relatively good performance on the first and second level questions than the third one, indicating that more investigation should be made on solving those questions that need reasoning.
\begin{table}[h]
	\begin{center}
		\begin{tabular}{lccc}
			\toprule 
			\bf State &\bf Total \# &\bf Right \# &\bf Accuracy\\
			\midrule
			All      & 100 &56&56.0\%\\
			\midrule
			First level & 30 & 18&60\%\\
			Second level &  30 &20&66.7\%\\
			Third level &  40 &18&45.0\%\\
			\bottomrule
		\end{tabular}
	\end{center}
	\caption{\label{tab5} Case analysis on RACE dataset.}
\end{table}

\subsection{Error Analysis}
In this part, a interesting example is shown in Figure \ref{fig:4}. Our model choose the right answer ``C". How can our model choose the right answer? Because the candidate ``C" has a bigest semantic overlap with passage. To improve that, we change candidate ``C" to {\em ``ellipsis in  talking"}, but the model can still choose the right answer. Even we change candidate ``C" to {\em``ellipses in  talking"}. From this experiment, we can see that our model can really handle part of high level semantic matching. 
Further more,  we change the question to {\em``The conversation is mainly  about   \_  ."}, the model still choose the candidate ``C" with a high probability. This reflects the problem that the deeplearning model is not sensitive to some special important information in passage and question. That phenomenon more or less approve the weakness of neural model. So we need to do more in-depth research to design some special component like CNN in dealing with picture to deal with text in reading comprehension.
\begin{figure}[tbp]
	\centering 
	\small
	\begin{tabular}{p{1\columnwidth}p{1\columnwidth}}
		\toprule
		\bf RACE Dataset \\ 
		\midrule
		\bf Passage \\ 
		As is known to all, in daily  {\color{blue}{conversation}} people often use simple words and simple sentences, especially {\color{blue}{elliptical}} sentences. Here is an interesting conversation between Mr Green and his good friend Mr Smith, a {\color{blue}fisherman}. Do you know what they are talking about? \\
		\midrule
		\bf Question  \\
		The text is mainly about   \_  . \\
		\midrule
		\bf Candidates  \\
		{\bf A}  how to catch  {\color{blue}{fish}}. \qquad\quad\quad
		{\bf B}  how to spend a Sunday \\
		{\bf C \em{\color{blue}{ellipsis}} in  {\color{blue}{conversations}} }
		\quad{\bf D}  {joy in  {\color{blue}{fishing}}}\\
		\bottomrule
	\end{tabular}
	\caption{Example of RACE dataset. The correct answer is depicted in bold face.}\label{fig:4} 
\end{figure} 
\section{Conclusion}\label{conclusion}
We propose the Convolutional Spatial Attention (CSA) model to tackle the machine reading comprehension with multiple-choice questions.
The proposed model could fully extract the mutual information among the passage, question, and candidates, to form the enriched representations using the modified attention mechanism that has trainable weights.
To summarize attention matrices from various sources, we propose to use convolutional operation to dynamically summarize the attention values within the different size of regions, which is beneficial to capture diverse features.
Experimental results show that the proposed CSA model could give substantial improvements over various state-of-the-art systems on both RACE and SemEval 2018 Task11 datasets.
Also, the ablation studies verify the effectiveness of several proposed components in our model, and analysis also shows that the CSA model is superior in solving relatively sophisticated questions (such as `why' or `how' questions).

\section{Acknowledgments}
We would like to thank all three anonymous reviewers for their constructive comments to improve our paper. This work was supported by the National Key R\&D Program of China via grant No. 2016YFC0800806.

\bibliography{aaai19}
\bibliographystyle{aaai}
\end{document}